\documentclass[11pt]{article}
\usepackage{latex/acl}
\usepackage{times}
\usepackage{latexsym}
\usepackage[T1]{fontenc}
\usepackage[utf8]{inputenc}
\usepackage{microtype}
\usepackage{inconsolata}
\usepackage{algorithm,algpseudocode}

\usepackage{amsmath}
\usepackage{amsfonts}
\usepackage{color,soul}
\usepackage{enumitem}
\usepackage{graphicx}
\usepackage{colortbl}
\usepackage{xcolor}
\usepackage{subfigure}
\usepackage{booktabs}
\usepackage[normalem]{ulem}
\useunder{\uline}{\ul}{}
\usepackage{efbox,graphicx}
\efboxsetup{linecolor=black,linewidth=2pt}

\usepackage{amsmath,amsfonts,bm}

\def\eqref#1{equation~\ref{#1}}

\def\1{\bm{1}}

\def\rmE{{\mathbf{E}}}

\def\rmG{{\mathbf{G}}}

\def\rmV{{\mathbf{V}}}

\def\ermE{{\textnormal{E}}}

\DeclareMathAlphabet{\mathsfit}{\encodingdefault}{\sfdefault}{m}{sl}
\SetMathAlphabet{\mathsfit}{bold}{\encodingdefault}{\sfdefault}{bx}{n}

\newcommand{\Pa}{\textrm{PA}_} 

\title{ACCESS : A Benchmark for \\ Abstract Causal Event Discovery and Reasoning}

\author{Vy Vo \quad Lizhen Qu \quad Tao Feng \quad Yuncheng Hua \quad Xiaoxi Kang\\
{\bf Songhai Fan \quad Tim Dwyer \quad Lay-Ki Soon \quad Gholamreza Haffari} \\
Monash University, Australia \\ 
\texttt{\{firstname.lastname\}@monash.edu}
}

\begin{document}
\maketitle
\begin{abstract}
  Identifying cause-and-effect relationships is critical to understanding real-world dynamics and ultimately causal reasoning. Existing methods for identifying event causality in NLP, including those based on Large Language Models (LLMs), exhibit difficulties in out-of-distribution settings due to the limited scale and heavy reliance on lexical cues within available benchmarks. Modern benchmarks, inspired by probabilistic causal inference, have attempted to construct causal graphs of events as a robust representation of causal knowledge, where \texttt{CRAB} \citep{romanou2023crab} is one such recent benchmark along this line. In this paper, we introduce \texttt{ACCESS}, a benchmark designed for discovery and reasoning over abstract causal events. Unlike existing resources, \texttt{ACCESS}  focuses on causality of everyday life events on the abstraction level. We propose a pipeline for identifying abstractions for event generalizations from \texttt{GLUCOSE} \citep{mostafazadeh-etal-2020-glucose}, a large-scale dataset of implicit commonsense causal knowledge, from which we subsequently extract $1,4$K causal pairs. Our experiments highlight the ongoing challenges of using statistical methods and/or LLMs for automatic abstraction identification and causal discovery in NLP. Nonetheless, we demonstrate that the abstract causal knowledge provided in \texttt{ACCESS} can be leveraged for enhancing QA reasoning performance in LLMs.
\end{abstract}

\section{Introduction}
\begin{figure*}[htb]
    \centering
    \includegraphics[width=\linewidth]{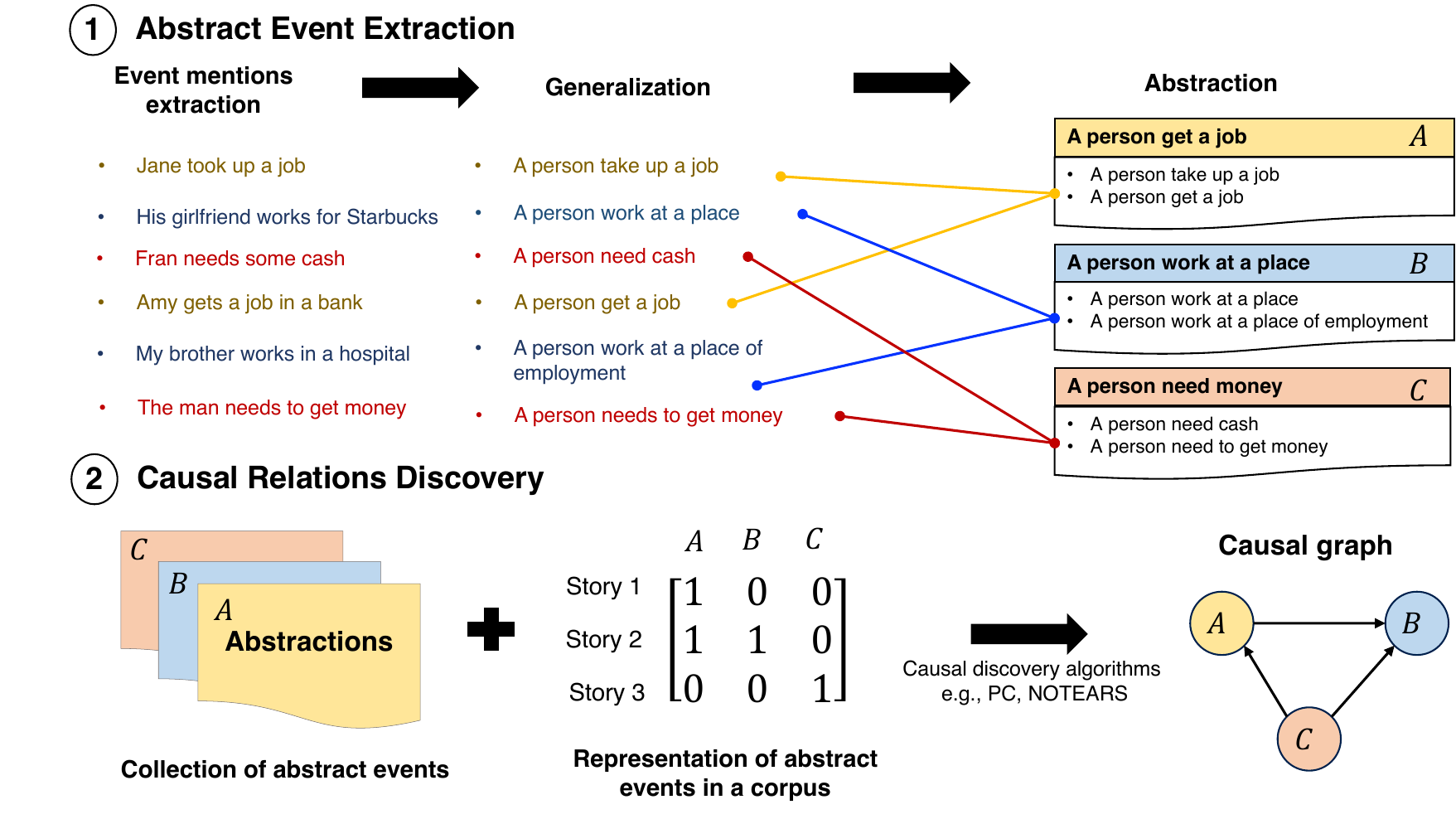}
    \caption{Pipeline of abstract causal event discovery. An event is viewed from three hierarchical levels: \textbf{mention} (realization in a specific text corpus), \textbf{generalization} (conceptualization of the event's components) 
    and \textbf{abstraction} (group of causally consistent generalizations). Given a collection of event mentions, Phase $1$ produces a collection of abstractions $A, B,C$ that are mapped back to the original corpus to construct a suitable representation in Phase $2$, such as a co-occurrence matrix. Causal discovery algorithms can then be employed to detect causal relations within the data, which may consider the contexts.}
    \label{fig:main}
\end{figure*}

Commonsense causal reasoning plays a vital role in developing a mental model of reality, where the ability to discover, explain and predict causal relations between events or forces in the environment is fundamental to human planning and control \citep{johnson2017mental, griffiths2017formalizing}. Cognitive science studies further suggest that event causality is critical to human understanding of narratives \citep{van1996children,fletcher1988causal,tillman2020children,sun2023event}, and story events with more causal relations tend to be better memorized than those with fewer relations \citep{graesser200310}. Humans are able to construct a causal mental model of events after reading a set of stories~\citep{zwaan1995construction}. For example, in Figure \ref{fig:main}, a reader would easily identify a causal relation between $e1:$ ``A person needs money.'' and $e2:$ ``A person gets a job.'' by \textit{abstracting} away concrete details, such as mentions of particular entities, grouping linguistic variations of the same meanings, and observing that $e1$ almost always leads to $e2$ in multiple stories, without explicit presence of lexical cues (e.g. \textit{because}) in text. Thus, this paper focuses on investigating to what extent LLMs can identify causal relations \textit{without relying on linguistic cues} and perform causal reasoning over commonsense knowledge on the abstraction level.

Prior works on \textit{causal relation extraction} heavily rely on linguistic cues, e.g. \textit{because of, by, due to}, to discern causal relations between event mentions and cause/effect text spans within a text~\cite{wolff2003models, mirza-tonelli-2014-analysis}. In contrast, statistical \textit{causal discovery} methods for event causality do not require linguistic cues but exploit statistical information of symbolic representations of events~\cite{pearl2018book}. As a result, those approaches are able to find causal relations even when they are not explicitly mentioned anywhere in texts. Therefore, there has been criticism regarding the susceptibility of these causal relation extraction models to exploit the linguistic cues to attain high performance without engaging in actual causal reasoning
\citep{yang2022survey,li2022counterfactual}. Ample of causal relation extraction datasets, including \texttt{TempEval-3} \cite{mirza2014annotating}, \texttt{CATENA} \cite{mirza2016catena}, \texttt{Causal-TimeBank} \cite{mirza-tonelli-2014-analysis}, \texttt{BECauSE} \cite{dunietz2015annotating, dunietz2017because} and \texttt{Event StoryLine Corpus} \cite{caselli2017event}, are not suitable for evaluating statistical causal discovery methods, because mentions of causal events that are semantically similar but expressed in different linguistic forms are not mapped into the same symbols. 

Humans identify grouping of semantic similar event mentions via \textit{abstraction}, which omits concrete spurious details. It has also been suggested that effective causal reasoning requires models to learn suitable abstract representation of the world \citep{girju2003automatic}. Abstraction of events can well leverage causality theories~\citep{pearl2009causality}, which provide a theoretical underpinning for causal reasoning and formal analysis of causal relations. Given a set of random variables, theoretically grounded causal discovery algorithms~\cite{vowels2022d} use statistics collected from a dataset to construct \textit{causal graphs}, which combine causal relations into logically coherent directed acyclic graphs (DAGs). In a causal graph, a node $v_i$ denotes a random variable, while an edge from $v_i$ to $v_j$ indicates $v_i$ is a direct cause of $v_j$. To the best of our knowledge, none of the existing datasets, such as \texttt{COPA} \citep{roemmele2011choice} and \texttt{MAVEN-ERE} \citep{wang-etal-2022-maven}, 
provide both abstraction of events grounded in a corpus and the corresponding causal graphs.

Existing knowledge graphs containing causal relations \cite{sap2019atomic,hwang2021comet,hassanzadeh2022knowledge,mbouadeu-etal-2023-evaluation}, cannot support evaluating both (1) event abstractions grounded in a corpus, and (2) construction of commonsense causal graphs from a collection of documents. 
The popular \texttt{ATOMIC} \citep{sap2019atomic} in particular is harvested by crowd-sourcing so that similar events in \texttt{ATOMIC} are not grouped together and there is also no associated corpus containing all relevant event mentions. As a result, it is challenging to recover the grounded contexts and map events to random variables in order to apply statistical causal discovery. Two additional resources for constructing causal graphs from a collection of documents are \texttt{CauseNet} \citep{heindorf2020causenet} and \texttt{CRAB} \cite{romanou2023crab}. 
The causal relations from \texttt{CauseNet} are explicitly mentioned in texts while causal semantics can exist beyond lexical mentions. \texttt{CRAB} generates causal graphs from events extracted from online news articles. Neither \texttt{CauseNet} or \texttt{CRAB} provides abstraction of events or grouping of semantically equivalent events with linguistic variations. Therefore, the resulting graphs over fine-grained events can explode in size, posing severe computational difficulties.  Another related resource is \texttt{GLUCOSE} \cite{mostafazadeh-etal-2020-glucose}, which translates natural language expressions of event mentions and their relations into generic inferential rules dependent on story contexts. In those rules, entities are mapped to generalized concepts and the same keywords are used for the same relations, such as \textit{causes}. Although \texttt{GLUCOSE} lacks grouping of similar abstract events and the inter-connection of relations into a logically coherent causal graph, our paper extends this dataset to construct abstract knowledge graphs of causality.

\paragraph{Contribution.}
We introduce \textbf{ACCESS}, a benchmark for \underline{\textbf{A}}bstra\underline{\textbf{C}}t \underline{\textbf{C}}ausal \underline{\textbf{E}}vent Di\underline{\textbf{S}}covery and Rea\underline{\textbf{S}}oning. We propose to explore event causality at the abstraction level as a more efficient representation of knowledge. 
We introduce a reusable pipeline (See Figure \ref{fig:main}) to curate causal relations from a large corpus of stories of daily life events. The resulting benchmark, \texttt{ACCESS}, is a graphical modelling of causal relations of $725$ event abstractions forming a base of abstract commonsense knowledge of causation. Our benchmark also includes annotations to evaluate each step of the pipeline. Using \texttt{ACCESS}, our experiments shed light on the ongoing challenges within the field. 

Firstly, the application of statistical structure learning algorithms for full graph discovery remains highly challenging. Secondly, solely relying on LLMs and automatic clustering proves insufficient for adequate event abstraction. Thirdly, LLMs still struggle with pairwise non-contextual causal discovery, indicating a gap in their possession of complete commonsense causal knowledge. Lastly, incorporating abstract commonsense knowledge through a causal graph enhances Question Answering (QA) reasoning tasks in LLMs up to $20\%$.

\section{Causal Event Abstraction}\label{sec:setup}
\begin{table*}[bt!]
\centering
\resizebox{\linewidth}{!}{%
\begin{tabular}{p{3cm} p{15cm}}
\toprule
\textbf{Terminology} & \textbf{Description} \\
\midrule
Event &  Any situation, state or action that happens, occurs or holds. 
An event consists of four basic components: \texttt{participant(s)}, \texttt{action/state}, \texttt{location} and \texttt{time}. \\
\midrule
Event sentence & An English sentence describing an event in daily life. An event sentence must contain the \texttt{participant(s)} and \texttt{action/state} components while \texttt{location} and \texttt{time} are the optional components and should not influence the judgment of the meaning of the sentence. \\ 
\midrule
Cluster & A group of sentences describing the same event. \\
\midrule
Topic & An event that is unique to a particular cluster and sufficiently abstract to be described by all event sentences in that cluster.  \\ 
\midrule
Topic sentence & The English sentence describing the topic of a particular cluster. \\ 
\midrule
Story & A description of a series of connected events. \\ 
\bottomrule
\end{tabular}
}
\caption{Terminologies of the \texttt{ACCESS} benchmark.}\label{tab:term}
\end{table*}

We follow the definition of events provided in TimeML \cite{pustejovsky2003timebank}, ECB+ Annotation Guidelines \cite{cybulska2014guidelines} and Event StoryLine Corpus \cite{caselli2017event}. An \textbf{event} refers to any situation or state that happens or holds, which consists of four basic components: \texttt{action/state, location, time} and \texttt{participant(s)}. We here consider \texttt{location} and \texttt{time} as optional; for instance, the sentence $he \ goes \ to \ sleep$ is sufficiently an event. Each component of an event is associated with a concept in an ontology.\footnote{Ontology refers to a collection of concepts and their relations within a domain \cite{gruber1993translation}.} A realization of a concept in the event is an  \textbf{event mention}. An \textbf{event abstraction} is a tuple $\langle$\texttt{action (state)/concept}, \texttt{participant/concept}, \texttt{time/concept}, \texttt{location/concept}$\rangle$ shared among all mentions of that event, where each component is either an entity or a concept at an appropriate abstraction level. An event abstraction is itself an event and can be identified by replacing every component in its representation by a more abstract concept in the ontology. For example, 
$His \ girlfriend$ \texttt{[person]} $works$ \texttt{[action]} $for \ Starbucks$ \texttt{[location]} $on \ the \ weekends$ \texttt{[time]}.

From another point of view, an event abstraction is a generalization of a \textbf{cluster} of event mentions that describe the same event. Two event mentions are \textbf{equivalent} if they are associated with the same event abstraction. An event abstraction is \textbf{causally consistent} w.r.t. a set of event mentions, if (1) none of its mention pairs at the semantic level contains a causal relationship, and (2) the semantics of all its mentions are either the cause or the effect of mentions in another event abstraction. Table \ref{tab:term} describes all the terms used in this paper and throughout the annotation process.

\paragraph{Definition of causation.} Based on the counterfactual theory of causation \citep{lewis2013counterfactuals}, an event $x$ is said to \textbf{cause} another event $y$ and event $y$ is said to be an \textbf{effect} of event $x$ if (1) event $y$ temporally follows event $x$ directly i.e., there are no intermediate events or if there is one, it must rarely occur, \underline{and} (2) event $y$ would not commonly occur if event $x$ did not occur. It is worth noting that unlike such datasets as \texttt{BECauSE} \citep{dunietz2017because} or \texttt{CauseNet} \citep{heindorf2020causenet} that consider causality between concepts, here causality is defined on the event (sentence) level, which takes into account the interaction of multiple participants. In statistical causality literature,  there exist $3$ causal structures of interest: \textit{confounder}, \textit{collider} and \textit{mediator}. For random variables $X, Y, Z$,

\begin{itemize}
    \item $Z$ is a called confounder if it causes both $X$ and $Y$, written as $X \leftarrow Z \rightarrow Y$;
    \item $Z$ is a collider when $Z$ is a common child of $X$ and $Y$ but $X$ and $Y$ themselves are not related, written as $X \rightarrow Z \leftarrow Y$;
    \item $Z$ is a mediator if there is a chain $X$ causes $Z$ and $Z$ causes $Y$, written as  $X \rightarrow Z \rightarrow Y$.
\end{itemize}

\paragraph{Quality criteria.}
We present the overarching criteria that guide our data construction process. These criteria aim to ensure that the event abstractions i.e., clusters of event mentions, in ACCESS achieve \textbf{causal consistency}:
\begin{enumerate} 
    \item Every cluster must be assigned with only one event abstraction. 
    
    \item All event mentions in each cluster must describe the same event and that event (abstraction) must be sufficiently abstract to cover all instances while being specific about the action taking place. 

    \item Every cluster must be in a cause-and-effect relation with at least one of the other clusters.


    \item If there exists a causal relation between events at one level, the causal relation must hold at its higher levels of abstraction in the hierarchy. For example, a causal relation between events at the \textit{mention} level must hold at the \textit{generalization} and \textit{abstraction} levels.

    \item A cluster $A$ is said to cause another cluster $B$ if \underline{at least one} event mentions in cluster $A$ causes any other event mentions in cluster $B$, according to the above cause-effect definition. 
\end{enumerate}

\begin{table*}[bt!]
\centering
\resizebox{\linewidth}{!}{%
\begin{tabular}{p{2.5cm} p{4.2cm}| p{3cm} p{6cm}}
\toprule
\multicolumn{2}{c}{\textbf{Cause event}} & \multicolumn{2}{c}{\textbf{Effect event}} \\
\midrule
Abstraction & Generalizations & Abstraction & Generalizations \\
\midrule
\textit{a person} & a person need money &  \textit{a person get} & a person take up a job  \\
\textit{need money} & a person need cash &  \textit{a job} & a person get a good job  \\
 & a person need to get money &  & a person get a job at a place  \\

\midrule
\textit{a person win} & a person win the contest & \textit{a person} & a person be celebrate an occasion  \\
 & a person win something &  \textit{celebrate} & a person have a celebration  \\
 
& a person end up winning &   & a person celebrate something  \\
\midrule
\textit{a person fall} & a person fall down & \textit{a person feel pain} & a person be in pain  \\
 & a person fall to the floor &   & a person experience pain in a body part  \\
 
& a person fall on the ground &  & a person 's body be in pain \\
\bottomrule
\end{tabular}
}
\caption{Examples of event causality on the abstraction and generalization level.}\label{tab:example}
\end{table*}

\section{The ACCESS Benchmark}
\texttt{ACCESS} provides a graphical modelling of the cause-and-effect relations among event abstractions, where every node in the causal graph represents an event abstraction in the causal relation between any two nodes is represented by an arrow going from the \textit{cause} event abstraction to the \textit{effect} event abstraction. There are $725$ abstractions or clusters, each of which on average contains $7$ instances, and in total associated with $9,513$ stories in the \texttt{GLUCOSE} dataset. The graph also contains diverse causal structures for causal inference, including confounding, mediation and collider \citep{pearl2009causality}. See Table \ref{tab:example} for examples of pairs of causal abstract events and Table \ref{tab:stats} for the descriptive statistics of \texttt{ACCESS}.

\begin{table}[h!]
    \centering
    \begin{tabular}{l|c c c r}
    \hline
    \multicolumn{5}{c}{\cellcolor[HTML]{C0C0C0}\textbf{Story corpus}}            \\ 
    \toprule
         Stories & & & & $9,513$  \\
         Events & & & & $4,708$  \\
         \hline
     \multicolumn{5}{c}{\cellcolor[HTML]{C0C0C0}\textbf{Causal graph}}            \\ \toprule
         Nodes (clusters / abstractions) & & & &  $725$  \\
         Edges (causal pairs) & & & &  $1,494$  \\ 
         Expected degree per node & & & &  $4$ \\
         Confounders & & & &  $149$  \\
         Mediators & & & &  $368$  \\
         Colliders & & & &  $3,956$  \\
         \bottomrule
    \end{tabular}
    \caption{General descriptive statistics of \texttt{ACCESS}.}
    \label{tab:stats}
\end{table}

Figure \ref{fig:main} illustrates our proposed pipeline for performing abstract causal event discovery and reasoning. The \texttt{ACCESS} dataset is constructed in the two phases: \textbf{Phase (1)} is to extract event abstractions from a collection of event mentions, by \textit{grouping mentions whose generalizations describing the same event} in a way that the resulting abstraction satisfies the above quality criteria. \textbf{Phase (2)} is to identify the causal relations among these event abstractions. Both phases entail an alternation between using automatic algorithms for extracting candidate clusters/causal pairs and crowd-sourcing for refinement and quality control. We briefly describe each phase in the following sections. See Appendix \ref{sup:annotation} for more details on our crowd-sourcing pipeline and task descriptions.

\subsection{Abstract Event Extraction}\label{sec:abstraction}
We now describe the process of curating these event mentions and extracting event abstractions. Our source of commonsense knowledge is \texttt{GLUCOSE} \citep{mostafazadeh-etal-2020-glucose}, a large-scale dataset of over $670$K stories with annotated causal relations. \texttt{GLUCOSE} also provides generalized inference rules mapped from specific statements, which correspond to our concept of event mentions.\footnote{For example, a specific statement $A \ neighbor \ knocked \ down \ my \ snowman$  is generalized into $Someone_{A} \ knocks \ down \ Something_{A}$.} We make use of the generalized expressions for our abstraction procedure and focus only on dimensions $1$ and $6$ of causal explanations: the direct effect. For simplicity, we will from now on refer to these generalizations as \textbf{events}. 

\paragraph{Automatic extraction.} Two or more event mentions must describe the same event to be clustered together. To describe the same event means they must be \textit{semantically related} or \textit{similar}. We initially apply standard text preprocessing and subsequently implement correlation clustering \citep{bansal2004correlation,charikar2005clustering} to automatically group events with shared semantics. We adopt an algorithm akin to the \texttt{PIVOT} algorithm \citep{fukunaga2019lp} that aims to maximize the semantic similarity of events in each cluster. We propose to measure semantic similarity by two metrics: \textit{cosine similarity} and \textit{paraphrasing likelihood}. The pairwise similarity of two expressions $x,y$ is given by
\begin{equation}\label{eq:sim}
   \mathcal{S}_{xy}= 0.5 \times \big[ f_{cos} (x, y) + f_{phr} (x, y)\big], 
\end{equation}
where $\mathcal{S}_{xy} \in [0,1]$, $f_{cos}$ returns the cosine similarity of the contextual embeddings of expressions $x,y$, and $f_{phr}$ returns the probability events $x,y$ are paraphrases. If $x,y$ are causally related, based on the annotations in \texttt{GLUCOSE},  $\mathcal{S}_{xy} = 0$. The contextual embeddings are obtained from the pre-trained \texttt{all-MiniLM-L6-v2} sentence Transformers \citep{reimers-2020-multilingual-sentence-bert} while the paraphrasing likelihood is obtained from the pre-trained adversarial paraphrase detector by \citet{nighojkar-licato-2021-improving}. 

Appendix \ref{sup:clustering} presents details of our clustering algorithm, which contains an ablation study against other popular clustering algorithms on unsupervised and supervised metrics to show that our \texttt{PIVOT} algorithm is preferable. In summary, the algorithm begins with a randomly chosen cause-effect pair of events as pivots. For each of these nodes, it finds the neighbors with which the similarity score exceeds $70\%$. The process is repeated for the remaining events until all events are clustered. Events that do not belong to any clusters are temporarily discarded. To ensure \textit{causal consistency}, we perform post-processing by splitting each cluster in a way that (1) no events in the same cluster are causally related, and (2) there exists either no or only one causal relations between any two clusters.

\paragraph{Human annotation.} We then utilize $10$ human annotators to assess the quality of cluster assignment as well as determine the abstract expression (or ``topic" in laymen term) for each cluster. This involves five key steps. First, the annotators are required to perform sub-clustering out of the clusters formed in the previous step. To strictly guarantee that events grouped together share the same semantics and maximize annotation consistency, we outline $11$ scenarios where word uses convey differences in meaning. Next, for each newly formed sub-cluster, they are also asked to identify the ``topic", which subsequently serves as an event abstraction. We then conduct three additional steps to resolve the disagreements in annotation as well as to handle the outlier events that are temporarily removed after the automatic procedure. Appendix \ref{sup:annotation_ph1} details this annotation process.

\subsection{Causal Relations Discovery}
This phase aims to identify the causal relations among the abstract events extracted from the previous phase, based on both non-contextual and contextual commonsense knowledge.  

\paragraph{Automatic causal discovery.}
To identify candidate causal pairs of event abstractions, we use a combination of existing annotated relations in \texttt{GLUCOSE} and statistical causal discovery methods. Regarding \texttt{GLUCOSE}, we determine the causal relation of two event abstractions (clusters) based on criterion $\#5$ in the above list of quality criteria. Regarding statistical causal discovery, we construct a dataset where each observation is a document or story in the \texttt{GLUCOSE} corpus and each feature records the counts of occurrences (or mentions) of a cluster in a story. On this co-occurrence data matrix, we run the well-known \texttt{PC} algorithm\footnote{a constraint-based structure learning method based on conditional independence tests.} \citep{spirtes2000causation} to obtain more causal candidates, using G-squared and Chi-squared tests at $p$-value of $0.01$. Note that we intentionally avoid using NLP models for event causality identification to avoid potential biases from their training data.  

\paragraph{Human annotation.} We proceed with human annotation on the union of the causal candidates from the above step. There are $3$ annotators participating in this task. They are asked to categorize each candidate causal pair $A$ and $B$ into three scenarios: $A$ \textit{causes} $B$, $B$ \textit{causes} $A$, or $A$ and $B$ have no relation. Initially, the workers are tasked with annotating the causal relations without considering contexts, that is to solely rely on their commonsense about the abstractions. Subsequently, we identify the causal pairs with no consensus from the three workers. We provide the story contexts in \texttt{GLUCOSE} associated with each of these pairs and ask them to reevaluate their annotations. Out of $2,862$ candidate pairs detected from \texttt{GLUCOSE}, $39.6\%$ of them are humanly annotated to be truly causal while that number is $61.5\%$ within \texttt{PC} candidates. The final relation of each pair is decided through majority voting. The inter-rate agreement score (Krippendorff’s $\alpha$) is $77.2\%$. See Appendix \ref{sup:annotation_ph2} for details.

\section{Experiments}\label{sec:exp}
In this section, we conduct empirical analyses to demonstrate how the \texttt{ACCESS} benchmark is used for evaluating (1) the effectiveness of automatic event abstraction and causal discovery approaches, and (2) how a causal structure assists reasoning models on causal QA tasks.  All experimental results are averaged over $5$ random running seeds. The codes and data for reproducing our experiments are published at \url{github.com/isVy08/ACCESS}. 

\subsection{Abstract Event Identification}\label{sec:abstraction_exp}For abstract causal discovery and reasoning, a practical question is how one can identify abstract events from real-world corpora where the ground-truth is unknown. 
Given the advances of LLMs, a promising approach to use LLMs to generate abstractions. In this experiment, we explore two approaches to automatically extract event abstraction with \texttt{GPT-4o-mini}, using Open AI's official API.\footnote{\url{platform.openai.com/} (accessed between 30 Sept. 2024 and 14 Oct. 2024).} We then use \texttt{ACCESS} as ground-truth to evaluate the quality of abstraction. 

\paragraph{Generate abstract events in a Single Step.}
We have \texttt{GPT-4o-mini} directly generate the generalized expressions. We extract $9,495$ event mentions from \texttt{GLUCOSE} and ask the model to generate two generalized versions for every instance, corresponding to the levels of \textit{generalization} (level $1$) and \textit{abstraction} (level $2$) described in Figure \ref{fig:main}. We then compare the generated abstractions with the ground-true ones provided by \texttt{GLUCOSE} and \texttt{ACCESS}. The model achieves the BLEU score of $0.520$. The prompt for this task can be found in Appendix \ref{sup:reasoning}.

\paragraph{Identify abstract events in Two Steps.}
In the second approach\footnote{In this experiment, we exclude duplicated expressions to reduce biases, resulting in $3,713$ and $4,248$ generalizations.}, we obtain the produced \textit{generalizations} by \texttt{GPT-4o-mini} from the above step, then run automatic clustering to find the abstractions, following the setup in Section \ref{sec:abstraction}. For all instances in every output cluster, we retrieve the ground-true clusters given by \texttt{ACCESS} and take the majority one as the predicted assignment. We measure the level of agreement between the predicted and the true assignment, using the Rand index \citep{steinley2004properties} and mutual information \citep{vinh2009information}. 

In Appendix \ref{sup:clustering}, Table \ref{tab:clustering_abs} provides detailed numerical results for various clustering algorithms in this experiment. In all cases, the agreement scores are well below $1.0$ (perfect agreement). This indicates  vanilla automatic clustering is inadequate in identifying useful abstractions. While choosing a good clustering algorithm remains important, we find that the quality of the input generalizations plays a more critical role in the performance. When we conduct the same experiments on the ground-true generalizations from \texttt{GLUCOSE}, all metrics are significantly improved by at least $28\%$. 

We further observe that on average, with generalizations from \texttt{GPT-4o-mini}, an output cluster has more than $40\%$ of its instances belonging to a different cluster from the predicted one, and based on the ground-truth, a cluster should be further divided into at least $2$ sub-clusters to be considered correct. We find that the issue is mainly due to the fact the model produces over-generalized expressions, causing the clustering algorithm to form bigger clusters. For example, the mentions \textit{Amanda feels excited} and \textit{He is scared} are both generalized to \textit{A person feel an emotion} while we consider \textit{be excited} or \textit{be scared} to refer to different states. Another example is the mention \textit{Tom works hard} being one-step generalized to be \textit{A person do something}, which arbitrarily can be applied to any expressions. This reveals the difficulty in controlling the granularity of abstractions using LLMs, which substantiates the necessity for the benchmarks on event generalization and abstraction.

\subsection{Pairwise Causal Discovery}\label{sec:CAUSAL}

We now describe how the data provided in \texttt{ACCESS} can be used for the causal discovery task. In the main text, we discuss the pairwise causal discovery task in LLMs. We examine how well LLMs can discern pairwise causal relations between two abstract events. Formally, given a pair of events $x$ and $y$, LLMs are asked to determine the relation between them by outputting one of the three possible relations: $x$ \textit{causes} $y$, $y$ \textit{causes} $x$, or \textit{no causal relation}. In addition to the $1,494$ causal relations in \texttt{ACCESS}, we also randomly generate $~1,000$ negative pairs to challenge the models. For our experiments, the LLMs used are \texttt{GPT-4o-mini}, \texttt{Llama3.2-3B-Instruct}, \texttt{Llama3.1-8B-Instruct} and \texttt{Llama2-chat-7B}\footnote{\url{llama.meta.com/} \& \url{huggingface.co/meta-llama/}}. The output from these models is post-processed to extract the final relation. The prompts can be found in Table \ref{tab:prompt_causal_discovery} of Appendix \ref{sup:reasoning}. 
 
The results are presented in Table \ref{tab:causal_discovery_results}, using Precision, Recall, and F1 score as evaluation metrics. We also report the performance of \texttt{random} choices and \texttt{majority} baseline, where the most frequent answer is selected for assessment based on the reference data.  LLMs achieve fairly humble accuracies, where \texttt{GPT-4o-mini} achieves the best performance, second to which is \texttt{Llama3.1-8B}. It is worth noting that the task is non-contextual since the goal is to assess the models' capability of intuitive or commonsense causal reasoning. Such intuition in humans is typically shaped by our observations and experiences from everyday life, enabling us to quickly identify scenarios where the causal relationships often hold. For example, we intuitively understand that speeding can frequently result in being the person being fined by the police. 

Appendix \ref{sup:causal_discovery} later demonstrates how the \texttt{ACCESS} pipeline facilitates the application of statistical structure learning algorithms. These methods are currently shown to under-perform on our benchmark, suggesting that there remains a large gap between theoretically grounded causal discovery and event causality identification research in NLP.   

\begin{table}[hbt!]
\centering
\resizebox{\linewidth}{!}{%
\begin{tabular}{l | r r r}
\hline
\multicolumn{4}{c}{\cellcolor[HTML]{C0C0C0}\textbf{Causal Discovery on \texttt{ACCESS}}}            \\ \toprule
                    & Precision $\uparrow$ & Recall $\uparrow$ & F1 $\uparrow$     \\ \toprule

\texttt{GPT-4o-mini}       & $\mathbf{0.705\pm.026}$    & $\mathbf{0.581\pm.028}$ & $\mathbf{0.559\pm.025}$ \\
\texttt{Llama3.2-3B} & $0.384\pm.015$    & $0.364\pm.006$ & $0.326\pm.007$ \\
\texttt{Llama3.1-8B} & $0.437\pm.006$    & $0.425\pm.006$ & $0.413\pm.006$ \\
\texttt{Llama2-7B} & $0.376\pm.006$    & $0.359\pm.006$ & $0.316\pm.007$ \\
 \midrule
\texttt{Random} & $0.340\pm.008$    & $0.330\pm.003$ & $0.330\pm.008$ \\
\texttt{Majority} & $0.114\pm.001$    & $0.333\pm.001$ & $0.170\pm.001$
 \\ 
\bottomrule
\end{tabular}%
}
\caption{Experiment results of causal discovery on \texttt{ACCESS} dataset. Precision, Recall, and F1 are computed under macro-average setting. \textbf{Bold} indicates best performance. $\uparrow$ Higher is better.}
\label{tab:causal_discovery_results}
\end{table}

\begin{table*}[!hbt]
\resizebox{\linewidth}{!}{%
\begin{tabular}{l|lr|lr|lr}
\hline
\multicolumn{7}{c}{\cellcolor[HTML]{C0C0C0}\textbf{QA Reasoning on \texttt{GLUCOSE}}}            \\ 
\toprule
 & \multicolumn{2}{c}{\textbf{Specific QA}}  & \multicolumn{2}{c}{\textbf{Specific QA+}}  & \multicolumn{2}{c}{\textbf{Abstract QA+}}  \\ 
 \midrule
zero-shot COT & Accuracy $\uparrow$   & F1 $\uparrow$     & Accuracy $\uparrow$   & F1 $\uparrow$    &  Accuracy $\uparrow$  & F1$ \uparrow$   \\
 \midrule
\texttt{GPT-4o-mini}      & $0.790\pm.008$                                    & $0.809\pm.006$           & $0.719\pm.049$           & $0.720\pm.019$           & $0.561\pm.013$           & $0.554\pm.013$           \\
\texttt{GPT-4o-mini} + CG & $\mathbf{0.894\pm.067}$                           & $\mathbf{0.887\pm.046}$  & $\mathbf{0.912\pm.012}$  & $\mathbf{0.887\pm.006}$  & $\mathbf{0.731\pm.047}$  & $\mathbf{0.686\pm.017}$  \\
\midrule
\texttt{Llama3.2-3B}      & $0.723\pm.006$ & $0.501\pm.010$  &      $0.696\pm.022$ & $0.487\pm.010$  & 
$0.631\pm.006$ & $0.524\pm.008$          \\
\texttt{Llama3.2-3B} + CG & $\mathbf{0.803\pm.002}$ & $\mathbf{0.561\pm.014}$ &   $\mathbf{0.754\pm.001}$ & $\mathbf{0.533\pm.002}$ & $\mathbf{0.725\pm.005}$ & $\mathbf{0.551\pm.004}$ \\
\midrule
\texttt{Llama3.1-8B}      & $0.883\pm.005$ & $0.428\pm.003$         & $0.833\pm.018$ & $0.418\pm.010$         & 
$0.794\pm.012$ & $0.421\pm.004$         \\
\texttt{Llama3.1-8B} + CG & $\mathbf{0.924\pm.008}$ & $\mathbf{0.556\pm.018}$ & $\mathbf{0.885\pm.002}$ & $\mathbf{0.521\pm.011}$ & $\mathbf{0.875\pm.001}$ & $\mathbf{0.524\pm.009}$ \\
\midrule
\texttt{Llama2-7B}        & $0.530\pm.013$ & $0.509\pm.012$  &     $0.480\pm.003$ & $0.553\pm.007$       &   
$0.483\pm.015$ & $0.537\pm.005$          \\
\texttt{Llama2-7B} + CG   & $\mathbf{0.681\pm.004}$ & $\mathbf{0.601\pm.009}$ & $\mathbf{0.635\pm.015}$ & $\mathbf{0.632\pm.010}$ & $\mathbf{0.692\pm.010}$ & $\mathbf{0.663\pm.009}$
\\
\bottomrule
\end{tabular}
}
\caption{Experiment results of multi-choice causal reasoning on \texttt{GLUCOSE-QA} dataset. \textbf{QA+} indicates the setting where the stories are paraphrased. \textbf{+CG} refers to the experiment that prompts the causal information from \texttt{ACCESS}. }
\label{tab:causal_reasoning_qa}
\end{table*}


\begin{table}[!h]
\centering
\resizebox{\columnwidth}{!}{%
\begin{tabular}{l|l r}
\hline
\multicolumn{3}{c}{\cellcolor[HTML]{C0C0C0}\textbf{Abstract QA+ on \texttt{GLUCOSE}}}            \\ \toprule
bi-level COT             & Accuracy $\uparrow$ &  F1 $\uparrow$     \\ 
\toprule
\texttt{Llama3.2-3B} & $0.722\pm.025$ & $0.378\pm.004$ \\
\texttt{Llama3.2-3B} + CG & $\mathbf{0.740\pm.025}$ & $\mathbf{0.410\pm.014}$ \\
\midrule
\texttt{Llama3.1-8B} & $0.753\pm.011$ & $0.430\pm.009$  \\
\texttt{Llama3.1-8B} + CG & $\mathbf{0.813\pm.022}$ & $\mathbf{0.493\pm.018}$ \\
\midrule
\texttt{Llama2-7B} & $0.503\pm.028$ & $0.516\pm.007$ \\
\texttt{Llama2-7B} + CG & $\mathbf{0.605\pm.018}$ & $\mathbf{0.623\pm.005}$ \\
 \bottomrule
\end{tabular}%
}
\caption{Experiment results of multi-choice Abstract QA+ with bi-level COT prompting. }
\label{tab:causal_reasoning_abstract_qa}
\end{table}


\subsection{Reasoning with Causal Graphs}\label{sec:QA}

We now study how the causal graphs in \texttt{ACCESS} can be used to assist models in QA reasoning tasks. In connection with Section \ref{sec:CAUSAL}, this can essentially be viewed as a contextual causal discovery task. We construct a causal QA dataset from \texttt{GLUCOSE}, which provides a set of stories with annotated causal relations between events at both the \textit{mention} and \textit{generalization} levels. For every story, we create two alternative multi-choice questions about the cause and effect. We extract the sentences in the story as candidate answers. The event appears in the annotated causal pair is thus considered a correct answer. Furthermore, in some cases, there can exist multiple causes that co-occur and lead to an effect and vice versa. To address this, we have a human expert review the data to identify additional correct causal events. The judgment of causality is based on the same criteria of the annotation process described in Appendix \ref{sup:annotation_ph2}. It is worth highlighting again that we focus on \textit{direct} causal relations, meaning that we do not consider events that result from/in some intermediate causes/effects. The resulting dataset contains $480$ questions, some of which have multiple correct answers. 

Since the LLMs have the tendency to exploit the textual cues, we additionally generate the paraphrases for each story while retaining the event choices in their original version. We label this extra setting as \textbf{QA+}. We also experiment with two variants of question design. In this first one, the cause/effect event of question occurs at the original mention level, whereas in the second one, the cause/effect event is transformed into its generalized version. For example, the question \textit{"What could be the cause of the event 'Amy gets a job in a bank'?"} is replaced into \textit{"The story describes an event where 'a person gets a job'. What could be the cause of the event?"}. We label the two variants as \textbf{Specific QA} and \textbf{Abstract QA} respectively. To perform the second task, ideally the model should be able to first perform abstract reasoning, that is to map the generalized cause/effect event to its corresponding mention in the context, prior to retrieving the correct causal pair. Examples of this \texttt{GLUCOSE-QA} dataset are provided in Appendix \ref{sup:reasoning}. 

\paragraph{How \texttt{ACCESS} provides abstract causal information.} To evaluate whether the causal abstract knowledge from \texttt{ACCESS} can help QA reasoning, we extend the above experiment by adding the causal relations between two relevant abstractions as an additional context in the prompt. In our experiments, the corresponding abstraction of an event mention can be retrieved directly from \texttt{ACCESS}. However, for an arbitrary QA dataset, this should be done via two subsequent steps: (1) perform abstraction of the event described in the target cause/effect and (2) map the output abstraction to at least one abstract event in \texttt{ACCESS} and retrieve the corresponding causal relations. Section \ref{sec:abstraction_exp} has described two possible approaches to step (1). 

\paragraph{Results.}
Given a story and a causal question, we prompt LLMs to generate the answer from a set of provided candidates. We first adopt zero-shot chain-of-thought (COT) \citep{kojima2022large} with the basic \textit{``let's think step by step"} prompting. Given the \texttt{Llama} models are open-sourced, we consider a bi-level COT dedicated to abstract QA tasks. In this approach, we provide a brief instruction on how to perform abstract causal reasoning, which entails two steps: the first one is for abstract reasoning, that is to identify the mention corresponding to the generalized cause/effect of question; the second step is for causal reasoning, that is to retrieve the corresponding effect/cause mentioned in the story context. See Appendix \ref{sup:reasoning} for prompts and qualitative examples.  

The evaluation metrics include Accuracy (which measures how often the model successfully retrieves at least one correct answer) and F1 score under weighted-average setting (which considers the alignment of all predicted choices). 
Tables \ref{tab:causal_reasoning_qa} and \ref{tab:causal_reasoning_abstract_qa} summarize our experiment results. Each setting introduces an increased level of difficulty in abstract reasoning. In the first task of Specific QA, the models can draw the answers directly from the raw context. Meanwhile, Specific QA+ tasks obscure away the linguistic cues, which the models are known to heavily exploit for prediction. Finally, Abstract QA+ is the most challenging, where the models are expected to concretize the abstract events before deriving the answers.  

The findings reveal that the inclusion of causal graphs significantly enhances the performance of LLMs across all experimental settings. Except \texttt{Llama2} whose performance is consistently poor, the performance of all models degrade on Abstract QA+ tasks, which indicates their struggle in reasoning over abstract causality. However, while we use \texttt{ACCESS} to provide the LLMs with the causal relations only between event abstractions, large improvements have been observed. Therefore, we hypothesize that the model may possess a subtle capacity to reason abstractly that needs proper activating. Compared to \texttt{GPT-4o-mini}, the \texttt{Llama} family are more prone to temporal and lexical biases, resulting in low F1 scores due to a higher number of negative selections. Concretely, the models select on average $1.9$-$2.9$ more answers than the actual ones across the QA tasks. With the additional causal information, the ratios are reduced to $1.8-2.6$. Bi-level COT unfortunately yields undesirable results, with a slight gain in accuracy in smaller models yet at a cost of reduced F1 scores due to increased over-prediction. This implies potential errors in some reasoning steps, but tracing and evaluating lines of reasoning in complex COT is an open challenge. Nevertheless, our experiments show that a simple zero-shot COT plus relevant abstract causal knowledge can greatly benefit the models. This presents a straightforward alternative strategy to enhance performance by leveraging external knowledge bases.

\section{Conclusion}
This paper introduces ACCESS, a benchmark for abstract causal event discovery and reasoning. We present a pipeline that combines automatic methods and human crowd-sourcing to extract $1,494$ causal relations among $725$ abstract events. We demonstrate that incorporating causal knowledge from our benchmark leads to improvements in QA reasoning tasks for LLMs. However, we also highlight challenges in automatic event abstraction identification and causal discovery, where in the latter, the popular statistical algorithms perform poorly in recovering our sub-graphs of fewer than $50$ nodes.  Our empirical evidence also suggests that LLMs are not ready to perform causal inference effectively due to the lack of effective acquisition of two critical sub-processes: abstract reasoning and causal discovery. This underscores the need for future research to equip the models with these essential skills for achieving true causal reasoning.

\section*{Limitations}
Our benchmark is built upon \texttt{GLUCOSE} \citep{mostafazadeh-etal-2020-glucose} whose scope is limited to everyday children's stories. Acknowledging this limitation, we propose a reproducible data construction pipeline applicable for curating diverse corpora of event causality.  Since \texttt{ACCESS} primarily addresses commonsense knowledge in real-world events, it is susceptible to biases regarding the judgement of semantic similarity and cause-and-effect relation of events. To mitigate this issue, our first effort is at every phase, to employ automatic methods alongside with human annotation, based on a set of objective definitions and criteria about events, abstractions and event causality. In the event abstraction phase, we specifically provide the annotators with a list of common scenarios (though non-exhaustive) indicating when the semantics of two expressions are considered similar of different to reduce potential biases.  Regarding the subjectivity in human causal judgment, while we focus on non-contextual causal commonsense knowledge, we leverage contextual signals in the original corpus whenever necessary to objectively guide the annotators' decisions on the causal relations. Due to the resource constraints, our causal graph is sparse and limited in size, which however still presents a challenge for statistical structure learning as well as LLMs on causal discovery tasks. One critical drawback in the experiment with statistical methods lies in the representation power of the co-occurrence matrix, which underscores the need for further research on representation learning of abstractions in language domain. As above, future works could also explore other resources to enlarge our causal graph and expand the coverage of real-world data. Such a causal graph could further be leveraged for causal inference according the engine described by Pearl \citep{pearl2009causality}, which seeks to answer causal queries across the three rungs of the Ladder of Causation i.e., associational (Rung 1), interventional (Rung 2), and counterfactual (Rung 3).

\section*{Ethics Statement}
To address potential misuse and uphold fairness and inclusivity, we discuss several ethical considerations for \texttt{ACCESS}. Firstly, it is crucial to clarify that the resources provided in this work are solely intended for research purposes. The narrative scenarios within \texttt{ACCESS} should not be utilized for insults, slander, or any other malicious purposes. Users are expected to adhere to the highest ethical standards, ensuring responsible and transparent use in line with ethical research practices. The creators of the dataset hold no responsibility for misuse or misinterpretation, and all necessary measures have been taken to respect privacy and ensure informed consent during the data collection process. Secondly, it is imperative to acknowledge the mental well-being of annotators during the data annotation process. Prior to data collection, this study underwent a thorough review and approval process by an internal review board. We require each annotator to take a break every two hours or whenever they feel uncomfortable.

\section*{Acknowledgment}
This material is based on research sponsored by DARPA under agreement number HR001122C0029. The U.S. Government is authorized to reproduce and distribute reprints for Governmental purposes notwithstanding any copyright notation thereon This work is also partially supported by the DARPA Assured Neuro Symbolic Learning and Reasoning (ANSR) program under award number FA8750-23-2-1016.

\bibliography{ref}
\clearpage

\appendix

\section{Related Work}
\paragraph{Theory of causation.} Extensive research into theories of causation spans various disciplines \citep{dalal2023calm} such as philosophy \citep{beebee2009oxford}, cognitive science \citep{waldmann2017oxford}, and probability and statistics \citep{pearl2009causality}. In this paper, we follow the counterfactual theory of causation \citep{lewis2013counterfactuals}, which entails three aspects: a \textit{relational} aspect (involving a cause and an effect components), a \textit{temporal} aspect (the cause precedes the effect), and a \textit{counterfactual} aspect (if the causing event had not taken place, the effect would not have occurred). 

\paragraph{Causal discovery in Statistics.} The task of causal discovery or structure learning is to recover the causal DAG using available observational or experimental data. It remains a challenging problem in statistics since the search space is super-exponential in the number of variables and the identifiability of the true DAG does not always exist. Causal discovery methods primarily fall into two categories: constraint-based and score-based approaches.  Constraint-based methods such as PC \cite{spirtes1991algorithm} and FCI \cite{spirtes2000causation} extract conditional independencies from the data distribution to detect edge existence and direction. Meanwhile, score-based methods search for model parameters in the DAG space by optimizing a scoring function \cite{chickering2002optimal,zheng2018dags,yu2019dag,bello2022dagma}.

\paragraph{Causal discovery in NLP.} Ample of work in NLP focuses on event causality identification (ECI), which identifies cause/effect spans from textual descriptions. ECI is commonly treated as a classification task that relies heavily on annotated data for supervised learning \cite{oh2013question, hashimoto2014toward, riaz2014recognizing, cheng2017classifying, gao2019modeling}, or at least partially for semi-supervised training \cite{zuo-etal-2021-improving, shen-etal-2022-event}. Machine learning models trained on mention-level annotations exploit event temporal links \cite{pustejovsky2003timebank, pustejovsky2006timebank} and/or lexical cues or semantics that signal causal information, including, but not limited to, prepositions e.g. \textit{because of, by, due to}, conjunctions/conjunctive adverbs e.g. \textit{because, since, therefore, as a result} or verb semantics \cite{wolff2003models, mirza-tonelli-2014-analysis} such as causation e.g. \textit{cause, force}. However, as these annotated benchmarks are relatively limited in scale, ECI models are prone to overfitting and tend to mishandle new and unseen cases \citep{zuo-etal-2021-improving, sun2023event}.

\paragraph{Causal discovery with LLMs.} Despite impressive language skills and breakthroughs in AI capabilities, large language models (LLMs) are reported to exhibit the same difficulty where they fail to perform causal inference in out-of-distribution settings when variable names and textual expressions used in the queries are different to those in the training set \citep{jin2023can,zevcevic2023causal}. On the other hand, whether LLMs can perform causal discovery is a controversial topic. \citet{kiciman2023causal} show that in medical and climate domains, LLMs can achieve competitive performance in determining pairwise causal relationships with accuracies up to $97\%$, yet heavily relying on prompt engineering with occasional inconsistencies. Meanwhile, full graph discovery in LLMs remains excessively challenging, though proper prompting could yield some potential. However, when evaluated on datasets of real-world events, \texttt{GPT-3} and \texttt{GPT-4} are consistently surpassed by small fine-tuned small pre-trained language models on ECI tasks \citep{gao2019modeling} while under-performing greatly on binary pairwise causality inference \citep{romanou2023crab}. Since LLMs are trained on massive volume of natural language texts, they excel in identifying causal event pairs but not non-causal ones, raises concerns regarding the memorization of event knowledge rather than generalization \citep{jacovi2023stop,gao2023chatgpt,romanou2023crab}. 

\paragraph{Abstract reasoning.} The human brain is equipped with a remarkable capability of abstract reasoning: thinking of concepts and generalizations of concrete entities that exist in infinity. Conceptualization glues separate pieces of experiences into a mental world that forms commonsense knowledge and allows us to function in the complex reality \cite{murphy2004big}. By the same logic, events sharing a physical mechanisms should exhibit the similar causal dynamics. For examples, $broken \ window$ and $shattered \ glass$ both refers to an effect resulting from a hard physical object hitting against a glassy surface. \texttt{GLUCOSE} \cite{mostafazadeh-etal-2020-glucose} is one large-scale annotated corpus that explicitly facilitates causal commonsense knowledge. The dataset captures 10 dimensions of causal explanations in story events covering a wider range of entities and contexts. \texttt{GLUCOSE} provides rich translations of specific expressions into generic inferential rules dependent on story contexts. \texttt{GLUCOSE} is thus a rich resource for exploiting abstract causal knowledge, which remains a promising yet unexplored avenue. In the common pursuit of abstract knowledge, \citet{he2022acquiring} build an event conceptualization pipeline on top of \texttt{ATOMIC} \cite{sap2019atomic}, a large-scale annotated commonsense knowledge graph, wherein the mentioned textual entities are replaced with the corresponding abstract concepts.

\section{Data Annotation Pipeline}\label{sup:annotation}

 
 

We recruit in total $13$ university students in Malaysia aged $20-30$. The total hours are  $329.7$, where the hourly rate is RM$20$ (Malaysian ringgit), which is higher than the minimum wage of RM$7.1$.  

As for the annotation guidelines, we translate the technical terminologies in Section \ref{sec:setup} into layman language comprehensible to human annotators.

\subsection{Abstract Event Extraction}\label{sup:annotation_ph1}
There are five steps in this annotation phase. Steps $1$ and $2$ are key to extracting abstract events, whereas Steps $3-5$ serve as post-processing to strengthen consistency among human annotators. 

\paragraph{Step 1: Sub-clustering.}
Each annotator is presented with a set of clusters generated from an automatic clustering algorithm. Each cluster contains multiple English sentences that describe events in daily life. Each word in every sentence is lemmatized to its base form so that the tense of the sentence does not influence the judgment of meaning. For every cluster, they are required to sub-group event sentences that are semantically similar or related together. There can exist clusters in which all sentences are related to one another; in this case no sub-clustering is needed. There can also be outlier events i.e., sentences that do not belong to any sub-clusters. For a sub-cluster to exist, it must contain at least two events. If an event cannot be sub-clustered, the annotator is to classify it as an outlier.	If a sentence is lexically or grammatically erroneous that makes it unjustifiable, the annotator is also asked to highlight and correct it whenever appropriate before clustering.

Two event sentences are considered \textit{semantically related} or \textit{similar}\footnote{We use ``='' to denote semantic similarity and ``$\ne$" to denote semantic dissimilarity between two events.} if they describe the same event, and the decision must not be affected by the information about \texttt{location} and \texttt{time}. We note there is a difference between a \texttt{state/action} actually taking place with the prospect of the \texttt{state/action} taking place. In particular, we outline $11$ scenarios where word uses convey differences in meaning.	

\begin{enumerate}
    \item single participant vs. group of participants e.g., \emph{a person be playing in the park} $\ne$ \emph{a person and another person be playing in the park.}
    
    \item affirmation vs. negation e.g., \emph{a person be asleep}  $\ne$ \emph{a person do not sleep.}

    \item present vs. future tense e.g., \emph{a person go to sleep} $\ne$ \emph{a person will go to sleep.}

    \item ability e.g., \emph{a person do not eat} $\ne$ \emph{a person cannot eat.}

    \item intention or desire e.g., \emph{a person do not eat} $\ne$ \emph{a person do not want to eat.}

    \item deduction or possibility e.g., \emph{it rain} $\ne$ \emph{it may rain.}. 
    
    \item obligation, advice or prohibition	e.g., \emph{a person do not eat} $\ne$ \emph{a person should not eat}.

    \item offers, effort or decision e.g., \emph{a person help another person} $\ne$ \emph{a person offer to assist another person}; \emph{a person go to the gym} $\ne$ \emph{a person decide to go to the gym.}

    \item location as object. In some cases, the object receives an action from the verb refers to a place or location e.g., \emph{a person clean a place}. Here \textit{room} is considered an (spatial) item being taken action on and similar to any other items such as cup or a table $\rightarrow$ \emph{a person clean a place} \textcolor{red}{=} \emph{a person clean something.}

    \item multiple actions. Some sentences describe two actions happening at the same time e.g., \emph{a person take something and leave}. In order to evaluate its meaning, one  must select one of them to the key action. The key action is the action that is described by most of other events in the same cluster. This means that if most of the other events are about \textit{someone leaving somewhere}, the \textit{leave} action should be focused instead of \textit{take} action. 

    \item continuous vs. simple tense. Some sentences describe actions in the continuous state e.g., \emph{a person be go home}. We ignore the continuous state of the action and consider them equivalent to the action described simple tense $\rightarrow$  \emph{a person be go home} \textcolor{red}{=} \emph{a person go home.}	
\end{enumerate}

\paragraph{Step 2 : Topic identification.}

In this step, the annotator asked to identify the topic for every cluster or sub-cluster formed. The topic must first be an event, therefore it must contain at least two components: \texttt{participant(s)} and \texttt{action}. The topic must be specific about the state or action that takes place. At the same time, the topic must be written in a way that makes it general or abstract enough to include all event sentences.	Whenever possible, it is acceptable to use the most representative event sentence in a cluster as the topic.	

\paragraph{Intermediate processing.} In Steps $1$ and $2$, we divide the collection of clusters into $7$ batches. Each of the batch contains $60$ clusters and 
two workers are asked to annotate one same batch of clusters. This results in one cluster having two annotation results. Subsequently, an algorithm is run to automatically unify the results from two annotators. For every cluster in the original data, the algorithm starts by randomly selecting an event as a centroid. It then forms a sub-cluster around the centroid that contains all other events that are considered by both annotators to be semantically related to the centroid. The topic assigned to that sub-cluster is presented in the format \texttt{TOPIC : [Text 1] / [Text 2]} where \texttt{[Text 1]} is the topic assigned to events in this sub-cluster by the first annotator and \texttt{[Text 2]} is the topic assigned to them by the second annotator. Repeat the process with the other events until all instances are processed. Thereafter, any event that is not assigned to any cluster will exist as a stand-alone instance and temporarily be considered an outlier. 

The next steps focus on resolving the disagreements from two annotation results, which includes \textbf{Topic alignment} and \textbf{Outliers processing}. We assume that a sub-cluster is properly annotated if it (1) contain at least $2$ instances and (2) no annotators consider that sub-cluster to be an outlier.

\paragraph{Step 3: Topic alignment.}
Every cluster is now annotated with two topics. If both topics describe the same event, the annotator is asked to choose either or the one more representative. Otherwise, choose the one that fits most of the sentences in the cluster. If the chosen topic is already assigned to some previous cluster, merge the current cluster into that cluster. If at least one of the topics is Outliers (i.e., at least one annotator considers the sub-cluster as Outliers), temporarily view them as Outliers. 

\paragraph{Step 4: Outliers processing.}
The annotator moves on to process the outliers. For any event that is assigned by only one of the previous annotators to be outliers while assigned by the other to be associated some existing sub-cluster, the annotator is asked to merge it into the assigned sub-cluster if the event can be represented by the topic of that sub-cluster; otherwise, keep it as an outlier. For any event that is agreed by both annotators to be an outlier, the current annotator is asked to re-examine it for possible assignment to any existing sub-cluster. The merging decision must be again based on the conditions described in Step $1$. Any remaining stand-alone instances are discarded. 

\paragraph{Step 5: Topic matching.} This step aims to correct for potential mis-clustering from the automatic procedure. We obtain the outlier events and attempt re-categorize them into the post-annotated clustering results from all above steps. For each outlier, we present the annotators with a set of candidate clusters to which adding the outlier would not violate causal consistency. We ask them to select one cluster with whose topic the outlier is most semantically similar. The rules to determine semantic similarity of a sentence pair follows from Step 1. It is possible that there is no topic that matches the outlier. If there is any topic that is a word-by-word exact match, that topic must be selected. We also add another rule that requires the annotators to select the topic with the same level of abstraction (generality) or concreteness (specificity) as the outlier event, since there are some topics that are abstract or concrete versions of other topics. More specifically, if the outlier is concrete but the concrete topic is not presented for selection, select the abstract topic. If the outlier is abstract but the abstract topic is not presented for selection, the concrete topic must \underline{not} be selected. 

\subsection{Causal Relations Discovery}\label{sup:annotation_ph2}
The annotator is tasked with evaluating candidate pairs of clusters to determine whether a cause-and-effect relationship exists between them, based on their respective topics. Since each cluster's topic represents an event abstraction, and in essence, an event itself, the decision on causal relation hinges on whether the two topics describe causally linked events. Based on the cause-effect definition in Section \ref{sec:setup}, we provide them with the following criteria to guide their decision about whether an event $A$ causes another event $B$:
\begin{enumerate}[leftmargin=5.5mm]
    \item a causal relation must be temporal, but a temporal relation is \underline{not} always causal;

    \item the action/state of $A$ directly leads to the action/state of $B$ i.e., there must be no intermediate events or if there is one, it should be extremely rare in real-world scenarios;
    
    \item an event $B$ would not occur if $A$ did not occur.
\end{enumerate}

Initially, the workers provide non-contextual annotations based solely on their commonsense understanding of the abstractions. A relation is deemed valid if the annotator can envision a plausible scenario in daily life where the situation occurs frequently, commonly, and is highly likely. If no such scenario comes to mind, the clusters are considered unrelated. In the subsequent step, we identify the highly disagreed pairs, where the three annotators each make distinct decisions regarding causality i.e., $A$ causes $B$, $B$ causes $A$, $A$ and $B$ are unrelated. For these pairs, workers are presented with contextual information from stories in \texttt{GLUCOSE} and asked to reconsider their decisions. The final determination of the relationship is made through majority voting.

\section{Clustering Algorithm}\label{sup:clustering}
Our clustering algorithm, named \texttt{PIVOT}, is inspired by the pivoting algorithm proposed in \citet{fukunaga2019lp}. The \texttt{PIVOT} algorithm first randomly selects a pair of cause-effect events and then, for each of them, find its most similar neighbors against a threshold of $70\%$. We repeat the process for the remaining event mentions, while excluding the previously assigned events. The initial results are passed to the following process to remove self-loop and bi-directions. We remove clusters with fewer than $10$ samples and maximum pairwise similarity is less than $50\%$. Each cluster can now be considered a node in a graph and we use \texttt{GLUCOSE} to recover the causal relations among them to construct a temporary causal graph. 

\paragraph{Ablation study.} The main motivation behind \texttt{PIVOT} algorithm is to ensure the initial graph is mostly acyclic while avoiding any sub-optimality produced from post-processing. To validate whether \texttt{PIVOT} is most effective in ensuring causal consistency, we conduct an ablation study against popular clustering algorithms, including \texttt{OPTICS} \cite{ankerst1999optics}, \texttt{LOUVAIN} \cite{blondel2008fast} and \texttt{LEIDEN} \cite{traag2019louvain} algorithms, where \texttt{LOUVAIN} and \texttt{LEIDEN} were proposed for community detection problems. The criteria for selecting these clustering algorithms include: 
(1) scalability to medium-to-large-sized data, (2) ability to accommodate custom affinity matrix and (3) high cluster homogeneity score. Table \ref{tab:clustering} further reports the quality of the algorithms under analysis, which shows that our \texttt{PIVOT} algorithm yields the most desirable performance.

\paragraph{Notations.} We use lower case letters (i.e., $v$) to denote single event, capital letters (i.e., $V$) for cluster of events, and blackboard bold letter (i.e., $\mathbb{V}$) for set of clusters. We let $\mathcal{D}$ denote the dataset of causal event mentions; $x \rightarrow y$ indicates event $x$ is a cause of event $y$; $x \leftarrow y$ indicates event $x$ is an effect of event $y$; $x \leftrightarrow y$ indicates $x$ and $y$ are causally related (either cause or effect). We also define the similarity between an event $y$ and cluster $V$ as the average of similarity between $y$ and every event $x$ in $V$
$$S_{yV} = \frac{1}{|V|} \sum_{x \in V} S_{xy},$$
where $S_{xy}$ is the similarity score between two events according to Eq. (\ref{eq:sim}).

\paragraph{Performance metrics.} In the following, we describe the unsupervised performance metrics to assess clustering algorithms in Table \ref{tab:clustering}. Given a set of clusters $\mathbb{C}$, let $\boldsymbol{A}$ be the matrix where $\boldsymbol{A}_{ij}$ is the number of events in cluster $C_i \in \mathbb{C} $ is the cause of any event in the cluster $C_j \in \mathbb{C}$. Recall that in this stage the causal relations between events are extracted from \texttt{GLUCOSE} dataset. A cluster $A$ is said to cause another cluster $B$ if \underline{at least one} event mentions in cluster $A$ causes any other event mentions in cluster $B$, according to the cause-effect definition in Section \ref{sec:setup}.

\begin{enumerate}
    \item \textit{Self-loop ratio}: Proportion of clusters in which the events are either cause or effect of each other. 
    $$\frac{1}{|\mathbb{C}|} \sum_{i=1}^{|\mathbb{C}|} \frac{\boldsymbol{A}_{ii}}{2 |C_i|}.$$

    \item \textit{Bi-directional ratio:} Proportion of cluster pairs that are both cause and effect of one another.  
    $$\frac{2}{|\mathbb{C}|^2 - |\mathbb{C}|} \sum_{i=1}^{|\mathbb{C}|-1} \sum_{j=i+1}^{|\mathbb{C}|} \frac{\min (\boldsymbol{A}_{ij}, \boldsymbol{A}_{ji})}{\max (\boldsymbol{A}_{ij}, \boldsymbol{A}_{ji})}.$$

    \item \textit{Silhouette coefficient \citep{rousseeuw1987silhouettes}:}  Measure of how similar an instance is to its own cluster (cohesion) compared to other clusters (separation). A high value indicates that the object is well matched to its own cluster and poorly matched to neighboring clusters.  
    $$\frac{1}{|\mathcal{D}|} \sum_{x \in \mathcal{D}} \frac{a_x - b_x}{1 - \min(a_x,b_x)},$$
    
    where $a_x$ is the mean similarity between event $x$ and all other events in the same cluster; $b_x$ is the mean similarity between event $x$ and all other events in the next nearest cluster.

    \item \textit{Homogeneity score:} Average pairwise similarity of events in a cluster.
    $$\frac{1}{|\mathbb{C}|} \sum_{i=1}^{|\mathbb{C}|} \frac{2}{|C_i|^2-|C_i|}\sum_{x,y \in C_i, x \ne y}S_{xy},$$
    
    where $S_{xy}$ is the similarity score between two events according to Eq. (\ref{eq:sim}).
\end{enumerate}

Table \ref{tab:clustering_abs} reports the numerical results for the experiment on Abstract Event Identification in Section \ref{sec:abstraction_exp}. For the supervised metrics, we refer readers to \href{https://scikit-learn.org/stable/modules/clustering.html}{\texttt{scikit-learn}'s} documentation for the precise formulations and implementations of \textit{Adjusted Rand Index} \citep{steinley2004properties} and \textit{Normalized Mutual Information} \citep{vinh2009information}. 

\begin{table*}[hbt!]
\centering
\begin{tabular}{l|r r r r}
\toprule
Metrics       & \textbf{LOUVAIN} & \textbf{LEIDEN} & \textbf{OPTICS} & \textbf{PIVOT}   \\
\midrule
Bi-directional ratio $\downarrow$               & $0.179$    & $0.162$   & $0.011$   & $\mathbf{0.004}$  \\
Self-loop ratio    $\downarrow$                & $0.252$    & $0.361$   & $0.007$   & $\mathbf{0.001}$  \\
Silhouette coefficient (Euclidean) $\uparrow$ & $-0.120$   & $-0.137$  & $-0.252$  & $\mathbf{-0.015}$ \\
Silhouette coefficient (Cosine) $\uparrow$    & $-0.234$   & $-0.262$  & $-0.392$  & $\mathbf{-0.036}$ \\
Homogeneity score $\uparrow$  & $0.506$  & $0.577$  & $0.810$   & $\mathbf{0.907}$ \\ 
\bottomrule
\end{tabular}
\caption{Evaluation of alternative clustering algorithms. \textbf{Bold} indicates best performance. $\uparrow$ Higher is better. $\downarrow$ Lower is better.} \label{tab:clustering}
\end{table*}

\begin{table*}[hbt!]
    \centering
\begin{tabular}{l|r r r r}
\toprule
Metrics       & \textbf{LOUVAIN} & \textbf{LEIDEN} & \textbf{OPTICS} & \textbf{PIVOT}$^{(*)}$   \\
\hline
\multicolumn{5}{c}{\cellcolor[HTML]{C0C0C0}\textbf{Generalizations from \texttt{GPT-4o-mini}}} \\ \toprule
Adjusted rand index $\uparrow$               & $0.016$    & $0.018$   & $0.001$   & $\mathbf{0.168}$  \\
Normalized mutual information $\uparrow$               & $0.450$    & $0.463$   & $0.384$   & $\mathbf{0.784}$  \\

\hline
\multicolumn{5}{c}{\cellcolor[HTML]{C0C0C0}\textbf{Generalizations from \texttt{GLUCOSE}}} \\ \toprule
Adjusted rand index $\uparrow$               & $0.042$    & $0.045$   & $0.011$   & $\mathbf{0.347}$  \\
Normalized mutual information $\uparrow$               & $0.635$    & $0.639$   & $0.699$   & $\mathbf{0.869}$  \\

\bottomrule
\end{tabular}
    \caption{Experimental results of using automatic clustering for identifying abstractions using generalizations by \texttt{ChatGPT} and human-annotated generalizations from \texttt{GLUCOSE}.  (*) In this experiment, we use the original implementation of the \texttt{PIVOT} algorithm in \citet{fukunaga2019lp}. \textbf{Bold} indicates best performance.}
    \label{tab:clustering_abs}
\end{table*}



\section{Statistical Causal Discovery}\label{sup:causal_discovery}

\paragraph{Background.} The causal relations among $n$ variables $X = [X_i]^{n}_{i=1}$ is characterized via a \textbf{structural causal model (SCM)} \citep{pearl2009causality} over the tuple $\langle U, X, f \rangle$ that, in its general form, consists of a sets of assignments
\begin{align*}
    X_i := f_i \big(\Pa{X_i}, U_i \big), \quad i = 1, \cdots, n,
\end{align*}
where $U_i$ is an exogenous variable assumed to be mutually independent with variables $\{U_1, \cdots, U_n\} \backslash U_i$. The functions $f = \left[f_1, \cdots, f_n \right]$ define a joint distribution $P(X)$ over the endogenous variables $X$, given a joint distribution over exogenous variables $P(U_1, \cdots, U_n)$. Each SCM induces a causal graph $\rmG$, which is often assumed to be a DAG. A directed graph $\rmG = (\rmV, \rmE)$ consists of a set of nodes $\rmV$ and an edge set $\ermE \subseteq \rmV^2$  of ordered pairs of
nodes with $(v, v) \notin \rmE$ for any $v \in \rmV$ (one without self-loops). 

For a pair of nodes $i,j$ with $(i,j) \in \ermE$, there is an arrow pointing from $i$ to $j$ and we write $i \rightarrow j$. Two nodes $i$ and $j$ are adjacent if either $(i,j) \in \rmE$ or $(j,i) \in \rmE$. If there is an arrow from $i$ to $j$ then $i$ is a parent of $j$ and $j$ is a child of $i$. Let $\Pa{X_i}$ denote the set of variables associated with parents of node $i$ in $\rmG$. The graph $\rmG$ of an SCM is obtained by creating one vertex for each $X_i$ and drawing directed edges from each parent $X_j \in \Pa{X_i}$ to $X_i$. We sometimes call the elements of $\Pa{X_i}$ the \textbf{direct causes} of $X_i$, and we call $X_i$ a \textbf{direct effect} of each of its direct causes. Importantly, these functions are to be read as assignments rather than as mathematical equations, and they should be viewed as modelling physical mechanisms inducing or generating every $X_i$ from variables $\Pa{X_i}$. 



\begin{figure*}[hbt!]
    \centering
    \includegraphics[width=\linewidth]{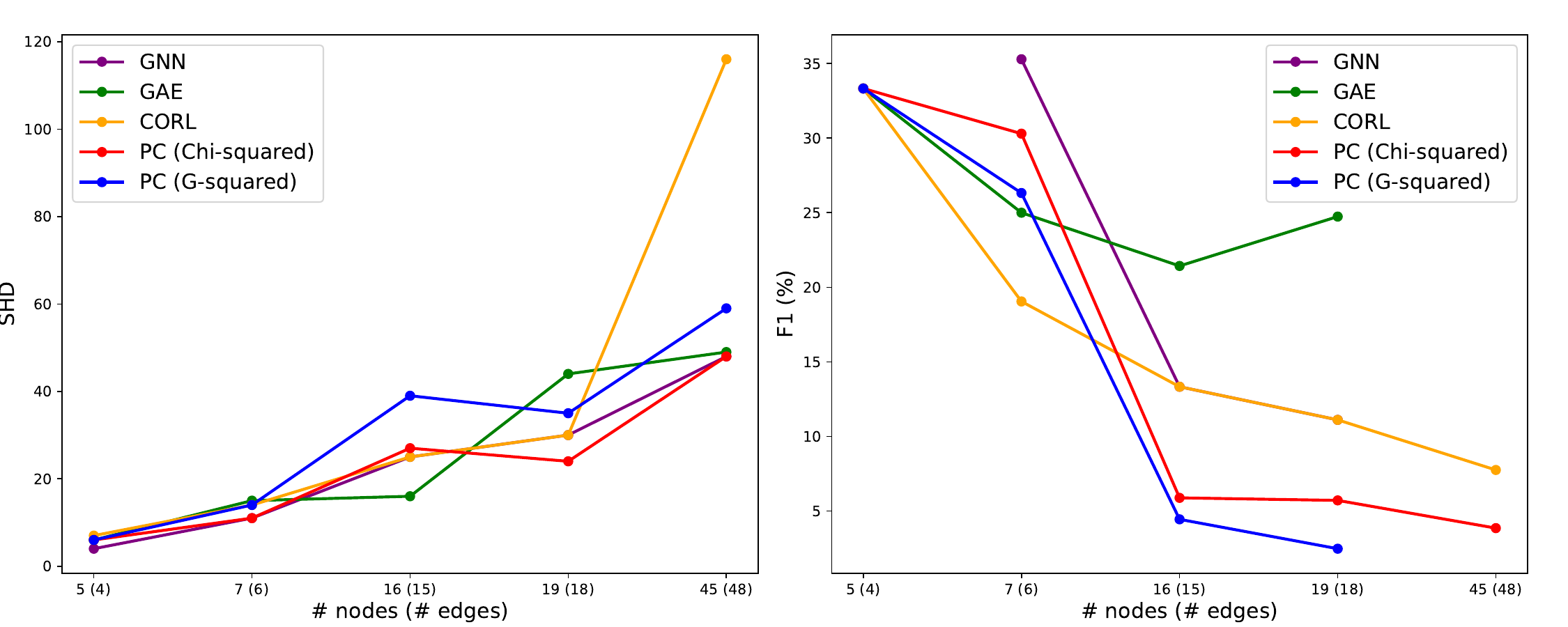}
    \caption{SHD \textbf{(left)} and F1 score \textbf{(right)} of estimated DAGs from statistical structure learning methods. \underline{Lower} SHD is better. \underline{Higher} F1 is better.}
    \label{fig:ssl}
\end{figure*}

\paragraph{Experiments.} We here discuss how \texttt{ACCESS} is used to assess to what extent the statistical structure learning methods is applicable to recover causal relations among event abstractions. As illustrated in Figure \ref{fig:main}, after extracting abstractions, one can build representations for abstract events in the original corpus and apply structure learning on top of such data for full graph discovery. A simple representation is the co-occurrence matrix size $(\# stories \times \# abstractions)$ where each entry takes a binary value indicating whether an abstraction has any of its mentions appearing in a story. This means each abstraction is now considered as a Bernoulli random variable and the task of causal discovery is to recover the underlying SCM where the structural functions are commonly non-convex. 

Due to the limited scalability of existing statistical algorithms, we resort to learning sub-graphs by setting thresholds to select nodes that appear frequently while ensuring that the true graph is acyclic. Specifically, our selected sub-graphs are composed of edges where both nodes are adjacent to at least one other node, and each node corresponds to an abstraction whose occurrences exceed a certain frequency threshold. In our experiment, we set thresholds for document frequency within  $\{25, 30, 35, 40, 45\}$, resulting in sub-graphs with $5, 7, 16, 19, 45$ nodes. The experiments are run on $5$ CPU cores.

We experiment with popular constraint-based and score-based algorithms. We select those that are scalable and capable of capturing non-linear causal relationships without relying on specific model forms such as additive noise. In this paper, we report the results for the following algorithms:

\begin{itemize}
    \item \texttt{PC} algorithm \citep{spirtes1991algorithm}: A classic approach based on conditional independence tests, for which we run two kinds of tests: Chi-squared and G-squared. 
    \item \texttt{DAG-GNN} \citep{yu2019dag}: DAG structure learning with graph neural networks.
    \item \texttt{GAE} \citep{ng2019graph}: This method utilizes gradient descent and graph auto-encoders to model non-linear causal relationships.
    \item \texttt{CORL} \citep{wang2021ordering}: A reinforcement learning-based algorithm with flexible score functions with enhanced efficiency.
\end{itemize}

Besides the above methods, we have also tested \texttt{NOTEARS} \citep{zheng2020learning}, a popular score-based algorithm and its more efficient variant \texttt{GOLEM} \citep{ng2020role}. However, they both unfortunately fail to recover any edges across all settings. To ensure consistency in implementation and evaluation, we utilize the standardized framework provided by \href{https://github.com/huawei-noah/trustworthyAI/tree/master/gcastle}{\texttt{gCastle}} \citep{zhang2021gcastle}. 
As for evaluation metrics, we report the structured Hamming distance (SHD), which quantifies the smallest number of edge additions, deletions, and reversals required to transform the recovered DAG into the true one. Additionally, we assess classification accuracy using the F1 score. Ideally, we aim for a lower normalized Hamming distance and a higher F1 score. Figure \ref{fig:ssl} reports the SHD and F1 score of the estimated DAGs from these methods. 

It is seen the methods achieve relatively low accuracy on our benchmark causal graphs, which are sparse. As the SHD scores are much higher than the graph size, these model tend to predict plenty of edges, most of which are incorrect due to the low F1 scores. Scalability remains a serious challenge to statistical structure learning. As the graph scales up to $45$ nodes, their performance further deteriorates significantly, where most of them of them even fail to recover any edges. It is noteworthy that the representation power of the input data also affects the causal discovery performance. It is very likely that the co-occurrence matrix is not sufficiently expressive to capture the causal knowledge. This motivates a dedicated line of research into abstract causal representation learning.

\section{GLUCOSE-QA Reasoning}  \label{sup:reasoning}
We here provide the prompts for LLMs in Tables \ref{tab:prompt_causal_discovery}-\ref{tab:prompt_abs}. Tables \ref{tab:examples_specific_qa}-\ref{tab:examples_cot_step} present illustrative examples of the responses from LLMs across our QA tasks.

\begin{table*}[!h]
    \centering
    \begin{tabular}{p{14cm}}
\toprule
\textbf{PROMPT: Pairwise Causal Discovery}\\
\midrule

\texttt{Given the two events:} \\ 

\texttt{event\_a: <Input the first event>}\\

\texttt{event\_b: <Input the second event>}\\

\texttt{Which cause-and-effect relationship is more likely between two events?} \\ 

\texttt{A. event\_a causes event\_b.}.  \\ 

\texttt{B. event\_b causes event\_a.} \\

\texttt{C. There are no cause-effect relation between two events.} \\

\texttt{Let’s work this out in a step by step way to be sure that we have the right answer. Then provide one final answer within the tags <answer>A or B or C</answer>.} \\

\bottomrule
\end{tabular}
\caption{Prompt for the pairwise causal discovery task.}
\label{tab:prompt_causal_discovery}
\end{table*}

\begin{table*}[!h]
    \centering
    \begin{tabular}{p{14cm}}
\toprule

\textbf{PROMPT: Multi-choice Answer Generation on Specific-QA (zero-shot COT)}\\
\midrule

\texttt{Given the following story: <Input story context>}.  \\ 

\texttt{What could be the <cause/effect> of the event <Input target effect/cause event>?} \\

\texttt{Choose one or more correct answers out of the following choices: <Input answer choices>.}

(*) \texttt{This information can help answer the question: A possible <cause/effect> of the event <Input effect/cause event abstraction> is <Input cause/effect event abstraction>.} 

\texttt{Let’s work this out in a step-by-step way to be sure that we have the right answer. Then provide your final answer beginning with `The correct answer(s):' followed by a list of the indices of the correct answers.} \\
\bottomrule
\end{tabular}
\caption{Prompt for the specific multi-choice answer generation on \texttt{GLUCOSE}. (*) This line is removed for the experiments that do not involve causal graphs.}
\label{tab:prompt_specific_qa}
\end{table*}

\begin{table*}[!h]
    \centering
    \begin{tabular}{p{14cm}}
\toprule
\textbf{PROMPT: Multi-choice Answer Generation on Abstract-QA (zero-shot COT)}\\
\midrule

\texttt{Given the following story: <Input story context>}.  \\ 

\texttt{The story describes an event where <Input generalization of target effect/cause event>. What could be the <cause/effect> of the event?} \\

\texttt{Choose one or more correct answers out of the following choices: <Input answer choices>.}

(*) \texttt{This information can help answer the question: A possible <cause/effect> of the event <Input effect/cause event abstraction> is <Input cause/effect event abstraction>.} 

\texttt{Let’s work this out in a step-by-step way to be sure that we have the right answer. Then provide your final answer beginning with `The correct answer(s):' followed by a list of the indices of the correct answers.} \\
\bottomrule
\\
\textbf{PROMPT: Multi-choice Answer Generation on Abstract-QA (bi-level COT)}\\
\midrule

\texttt{Given the following story: <Input story context>}.  \\ 

\texttt{The story describes an event where <Input generalization of target effect/cause event>. What could be the <cause/effect> of the event?} \\

\texttt{Choose one or more correct answers out of the following choices: <Input answer choices>.}

(*) \texttt{This information can help answer the question: A possible <cause/effect> of the event <Input effect/cause event abstraction> is <Input cause/effect event abstraction>.} 

\texttt{The event [Input generalization of target effect/cause event> is described by one of the sentences in the story context. First identify that part of the story. Then retrieve the event mentioned in the story that is a corresponding cause/effect.".} 

\texttt{Let’s work this out in a step-by-step way to be sure that we have the right answer. Then provide your final answer beginning with `The correct answer(s):' followed by a list of the indices of the correct answers.} \\
\bottomrule
\end{tabular}
\caption{Prompts for the abstract multi-choice answer generation on \texttt{GLUCOSE}. (*) This line is removed for the experiments that do not involve causal graphs.}
\label{tab:prompt_abstract_qa}
\end{table*}

\begin{table*}[]
    \centering
    \begin{tabular}{p{14cm}}
\toprule
\textbf{PROMPT: Abstract Event Identification}\\
\midrule
\texttt{We need to convert the input sentence into a more general expression. The conversion consists of three steps.} \\

\texttt{First, identifying: identify entities and verb words.} \\ 
\texttt{Second, conversion: convert the entities with more generic words and transform the verb words into the base form.} \\
\texttt{Third, further conversion: convert the sentence into a more general expression.} \\
\texttt{Note: The generic expressions used in the conversion are placeholders for the specific details in the original sentence.}

---------------------------------------------------------- \\ 
\texttt{The following is a conversion example.} \\

\texttt{Original Sentence:} \textit{John went to buy a new collar for his dog.}

\texttt{1. Identifying:}
\begin{itemize}[noitemsep]
    \item \texttt{Person:} \textit{John}
    \item \texttt{Action:} \textit{went, buy}
    \item \texttt{Object:} \textit{a new collar}
    \item \texttt{Possession:} \textit{his dog}
\end{itemize} 
        
\texttt{2. Conversion:} \textit{a person go to buy another thing for something} \\

\texttt{3. Further Conversion:} \textit{a person buy something to do something}

---------------------------------------------------------- \\
        
\texttt{The following is another example.}

\texttt{Original Sentence:} \textit{John drives near the woman.} 

\texttt{1. Identifying:}
\begin{itemize}[noitemsep]
    \item \texttt{Person:} \textit{John}
    \item \texttt{Action:} \textit{drives}
    \item \texttt{Object:} \textit{the woman}
    \item \texttt{Preposition:} \textit{near}
\end{itemize} 
        
\texttt{2. Conversion:} \textit{a person see another person} \\

\texttt{3. Further Conversion:} \textit{a person see another person}

---------------------------------------------------------- \\
        


        



\texttt{Now we have a test instance. Please refer to the task instruction and the above examples to do the conversion.} \\
\texttt{The input sentence is: <Input event mention>}.\\
\texttt{Please convert the sentence into a more general expression following the above-mentioned three steps.}\\ 
\bottomrule

    \end{tabular}
    \caption{Prompt for the abstract event identification task.}
    \label{tab:prompt_abs}
\end{table*}

\begin{table*}[hbt!]
\centering
\resizebox{\linewidth}{!}{%
\begin{tabular}{l|p{12cm}}
\toprule
Story & In a store, two women were arguing, and Howard wanted to intervene. He attempted to get them to stop talking, but it didn't work. So, he stepped in between them, which caused them to cease their fighting. \\
\midrule
Specific Question & What could be the cause of the event \textit{`howard wants to help the women'}? \\
\midrule
Abstract Question & The question describes an event where \textit{`a person hears something in a place'}. What could be the effect of the event?\\
\midrule
Choices & \begin{tabular}[c]{@{}l@{}}0: "Two women fights each other.",\\ 1: "He went in between them.",\\ 2: "Two women were fighting in a store.",\\ 3: "They stopped.",\\ 4: "Howard wanted to help."\\ 5: "He tried telling them to stop but it did not work."\end{tabular} \\ \midrule
Causal Graph (CG) & \textit{a person have a fight with another person} $\rightarrow$ \textit{a person want to stop another person} \\ \midrule
Correct Answers & 0, 2 \\ \midrule
\texttt{GPT-4o-mini} Answers & 2, 4 \\ \midrule
\texttt{GPT-4o-mini} Answers w/ CG & 0, 2 \\ \midrule
\texttt{Llama3.1-8B} Answers & 0, 1 \\ \midrule
\texttt{Llama3.1-8B} Answers w/ CG & 0, 2, 4 \\ 
\bottomrule
\end{tabular}%
}
\caption{Examples of multi-choice Specific-QA reasoning in \texttt{GPT-4o-mini} and \texttt{Llama3.1-8B}.}
\label{tab:examples_specific_qa}
\end{table*}

\begin{table*}[hbt!]
\centering
\resizebox{\linewidth}{!}{%
\begin{tabular}{l|p{12cm}}
\toprule
Story & His cousins were scheduled to visit later that day, so his mom had him clean in the morning, shop for groceries in the afternoon, and get ready in the evening. Eventually, his cousins arrived at his house. \\
\midrule
Abstract Question & The question describes an event where \textit{`a person are coming to a place (that is another person house)'}. What could be the effect of the event?\\
\midrule
Choices & \begin{tabular}[c]{@{}l@{}}0: "His cousins were coming later too his house.",\\ 1: "He get groceries in the afternoon.",\\ 2: "His mom made him clean all morning.",\\ 3: "His cousins came to his house.",\\ 4: "He get ready in the evening."\end{tabular} \\ \midrule
Causal Graph (CG) & \textit{a person come to another person 's place} $\rightarrow$ \textit{a person clean something} \\ \midrule
Correct Answers & 1, 2, 4 \\ \midrule
\texttt{GPT-4o-mini} Answers & 0, 3 \\ \midrule
\texttt{GPT-4o-mini} Answers w/ CG & 0, 2 \\ \midrule
\texttt{Llama3.2-3B} Answers & 0, 3 \\ \midrule
\texttt{Llama3.2-3B} Answers w/ CG & 1, 3 \\ 
\bottomrule
\end{tabular}%
}
\caption{Examples of multi-choice Abstract-QA reasoning in \texttt{GPT-4o-mini} and \texttt{Llama3.2-3B}.}
\label{tab:examples_abstract_qa_1}
\end{table*}

\begin{table*}[hbt!]
\centering
\resizebox{\linewidth}{!}{%
\begin{tabular}{l|p{12cm}}
\toprule
Story & Felix wanted to visit Disney World. One day, he won two tickets and invited his friend Alissa. However, Alissa disliked Disney, so Felix ended up going by himself. \\
\midrule
Abstract Question & The question describes an event where \textit{`a person invited another person'}. What could be the cause of the event?\\
\midrule
Choices & \begin{tabular}[c]{@{}l@{}}0: "Alissa hated disney.",\\ 1: "Felix wanted to go to disney world.",\\ 2: "One day he won two tickets for entry.",\\ 3: "He invited his friend Alissa.",\\ 4: "He ended up going alone."\end{tabular} \\ \midrule
Causal Graph (CG) & \textit{a person want to go to a place} $\rightarrow$ \textit{a person give another person an invitation to a place} \\ \midrule
Correct Answers & 1, 2 \\ \midrule
\texttt{GPT-4o-mini} Answers & 0, 1, 3 \\ \midrule
\texttt{GPT-4o-mini} Answers w/ CG & 1, 2 \\ \midrule
\texttt{Llama2-7B} Answers & 1, 2 \\ \midrule
\texttt{Llama2-7B} Answers w/ CG & 1, 2 \\ 
\bottomrule
\end{tabular}%
}
\caption{Examples of multi-choice Abstract-QA reasoning in \texttt{GPT-4o-mini} and \texttt{Llama2-7B}.}
\label{tab:examples_abstract_qa_2}
\end{table*}
\begin{table*}[hbt!]
\centering
\resizebox{\linewidth}{!}{%
\begin{tabular}{l|p{12cm}}
\toprule
Story & He wanted toast, so he got some bread and put it in the toaster. When it popped out and landed on the floor, he ate it anyway. \\
\midrule
Abstract Question & The question describes an event where \textit{`a person got another thing (that is an ingredient in another thing'}. What could be the cause of the event?\\
\midrule
Choices & \begin{tabular}[c]{@{}l@{}}0: "He ate it anyway.",\\ 1: "He put it in the toaster.",\\ 2: "He got some bread.",\\ 3: "It shot out of the toaster and onto the floor.",\\ 4: "He was making toast."\end{tabular} \\ \midrule
Correct Answers & 4 \\ 
\midrule
\texttt{Llama3.2-3B} Answers (zero-shot) & 1, 2 \\ 
\texttt{Llama3.2-3B} Answers & 1, 4 \\ 
\texttt{Llama3.2-3B} Answers + CG & 1, 4 \\ 
\midrule
\texttt{Llama3.1-8B} Answers(zero-shot) & 1, 3 \\ 
\texttt{Llama3.1-8B} Answers  & 2 \\ 
\texttt{Llama3.1-8B} Answers + CG & 4 \\ 
\midrule
\texttt{Llama2-7B} Answers (zero-shot) & 1, 2 \\ 
\texttt{Llama2-7B} Answers & 4 \\ 
\texttt{Llama2-7B} Answers + CG & 1, 4 \\ 
\bottomrule
\end{tabular}%
}
\caption{Examples of multi-choice Abstract-QA reasoning with bi-level COT.}
\label{tab:examples_cot}
\end{table*}

\begin{table*}[hbt!]
\centering
\resizebox{\linewidth}{!}{%
\begin{tabular}{l|p{12cm}}
\toprule
Story & He wanted toast, so he got some bread and put it in the toaster. When it popped out and landed on the floor, he ate it anyway. \\
\midrule
Abstract Question & The question describes an event where \textit{`a person got another thing (that is an ingredient in another thing'}. What could be the cause of the event?\\
\midrule
Choices & \begin{tabular}[c]{@{}l@{}}0: "He ate it anyway.",\\ 1: "He put it in the toaster.",\\ 2: "He got some bread.",\\ 3: "It shot out of the toaster and onto the floor.",\\ 4: "He was making toast."\end{tabular} \\ \midrule
Prompt & The event \textit{`a person got another thing (that is an ingredient in another thing'} is described by one of the sentences in the story context. First identify that part
of the story. Then retrieve the event mentioned in the story that is a
corresponding cause/effect.\\ 
\midrule
Correct Answers & 4 \\ 
\midrule
Rationale & To solve this question, let's break down the steps: \\
& 1. Identify the sentence describing `another thing which is an ingredient in another thing': - The relevant sentence from the given text is not explicitly stated but rather implied through analysis. In this case, what seems like `he made breakfast food'. In other words - Toaster = Breakfast Food; Bread + Toast = Ingredients for the meal at hand.  Therefore the first item on our consideration has been identified as number four because if you are going into kitchen then most likely, your objective would be about cooking something.\\ 
& 2. Now find any events occurring after items were used during said activity using index numbers provided against them starting their respective sequence no. This means looking back towards end, start at \#5 because there will definitely exist only those indexed things found before each index respectively being associated w/ this particular action taking place afterward within same timeframe i.e., related actions happening after certain things are prepared here. Considering these points: We look forward now...4 provides our initial object `the ingredients'. This makes us say - That must come after getting things ready prior than eating. Therefore we also know - getting stuff does indeed precede putting things inside the device whose result gets eaten later.\\ 
& So while option zero fits too since all meals need ending act (eating) coming second to doing preparatory tasks involving having desired objects (`bread').. But finally- lets verify... As shown below, the correct answer(s): 4, 1\\ 
\bottomrule
\end{tabular}%
}
\caption{Example of multi-choice Abstract-QA reasoning with bi-level COT in \texttt{Llama3.2-3B}.}
\label{tab:examples_cot_step}
\end{table*}

\end{document}